%% file: main.tex
\documentclass[10pt,twocolumn,letterpaper]{article}

\usepackage{cvpr}              
\usepackage{csquotes}
\usepackage{comment}
\usepackage{mdframed} 
\usepackage{adjustbox}
\usepackage{tikz}
\usetikzlibrary{arrows.meta,positioning,fit,calc}
\input{preamble}
\definecolor{cvprblue}{rgb}{0.21,0.49,0.74}
\usepackage[pagebackref,breaklinks,colorlinks,allcolors=cvprblue]{hyperref}

\def\paperID{37} 
\def\confName{EarthVision}
\def\confYear{2026}

\title{Open-SAT: LLM-Guided Query Embedding Refinement for Open-Vocabulary Object Retrieval in Satellite Imagery}

\author{
Md Adnan Arefeen\thanks{Work performed while employed at NEC Laboratories America.}\\
North South University\\
Dhaka, Bangladesh\\
{\tt\small adnan.arefeen@northsouth.edu}
\and
Biplob Debnath, Ravi K. Rajendran\\ Murugan Sankaradas\footnotemark[1], Srimat T. Chakradhar\\
NEC Laboratories America\\
Princeton, NJ, USA\\
{\tt\small \{biplob, rarajendran, murugs, chak\}@nec-labs.com}
}

\begin{document}
\maketitle
\input{texFiles/abstract}

\input{texFiles/intro}
\input{texFiles/method}

\input{texFiles/results_earthvision}

\input{texFiles/system}
\input{texFiles/relatedwork}

\input{texFiles/conclusion}
{
    \small
    \bibliographystyle{ieeenat_fullname}
    \bibliography{main}
}


\end{document}

%% file: texFiles/abstract.tex
\begin{abstract}

In satellite applications, user queries often take the form of open-ended natural language, extending beyond a fixed set of predefined categories. This open-vocabulary nature poses significant challenges for retrieving relevant image tiles, as the retrieval system must generalize to a wide range of unseen objects and concepts. While vision-language models (VLMs) such as CLIP are widely used for text-image retrieval, even fine-tuned variants often struggle to accurately align such queries with satellite imagery. To address this, we propose \textbf{Open-SAT}, a training-free query embedding refinement algorithm that operates at inference time to improve alignment between user queries and satellite image content. Open-SAT uses VLMs to compute embeddings for image tiles, which are stored in a vector database for efficient retrieval. At query time, it leverages Large Language Models (LLMs) to refine the text embeddings by incorporating contextual information about objects of interest and their surroundings. A threshold-free retrieval mechanism further enhances accuracy and efficiency.
Experimental results in three public benchmarks demonstrate that Open-SAT improves the F1 score by up to 16.04\%, while retrieving a comparable number of image tiles. These results demonstrate the effectiveness of Open-SAT in open-vocabulary satellite image retrieval, leveraging LLM guidance without the need for additional training or supervision.

\end{abstract}

%% file: texFiles/intro.tex
\section{Introduction}

Satellites play a vital role in monitoring the Earth's surface, generating high-resolution, multimodal imagery that supports a wide range of applications. These images are captured at varying spatial resolutions, ranging from low resolution ($> 30$ m/pixel) to very high resolution ($< 1$ m/pixel). Modern satellite systems also provide frequent revisit rates, enabling high-temporal-resolution monitoring of environmental and anthropogenic changes.

The availability of free satellite imagery from sources such as Landsat, Sentinel, and MODIS has significantly enhanced access to Earth observation data. Platforms like NASA's Earthdata Search and ESA's Copernicus Data Space Ecosystem provide access to extensive satellite datasets, including both historical and near real-time imagery. These resources empower researchers, policymakers, and businesses to utilize satellite data for applications such as environmental monitoring, urban planning, risk assessment, and disaster management.  

In this paper, we address the challenge of handling open-vocabulary queries related to diverse objects in satellite images, enabling a broad range of user applications. For example, a user might upload a satellite image and ask questions such as, \enquote{How many residential houses have solar panels installed?} or \enquote{Find construction sites}. They may also inquire about specific objects, such as buildings or vegetation types.  

\begin{figure*}[h]
    \centering
    \begin{subfigure}{0.3\linewidth}
        \centering
        \includegraphics[width=\linewidth]{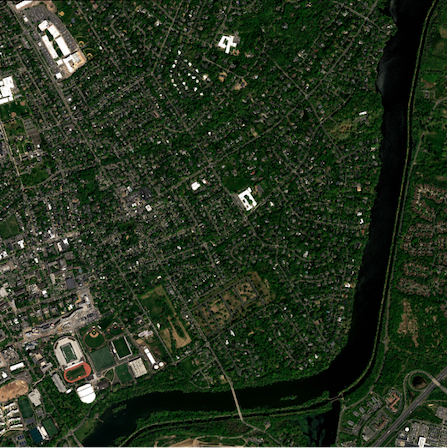} 
        \caption{}
    \end{subfigure}
    \hfill 
    \begin{subfigure}{0.3\linewidth}
        \centering
        \includegraphics[width=\linewidth]{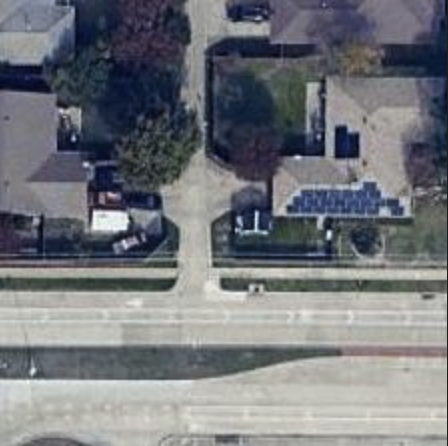} 
        \caption{}
    \end{subfigure}
    \hfill 
    \begin{subfigure}{0.3\linewidth}
        \centering
        \includegraphics[width=\linewidth]{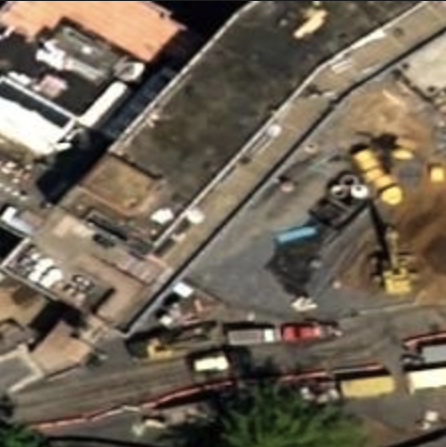} 
        \caption{}
    \end{subfigure}
    \caption{(a) A satellite image of Princeton area covering $10km\times7km$, (b) a tile containing solar panel, and (c) a tile containing construction zone.}
    \label{fig:example-princeton}
\end{figure*}

Satellite images often span large geographic areas, rendering objects like cars, buildings, or trees as just a few pixels. Detecting these small objects with accuracy requires specialized algorithms capable of managing substantial scale variations while preserving precision at low resolutions. Recent research has increasingly focused on small-object detection, aiming to extract insights based on user queries~\citep{biswas2024domain, chen2024micpl, yang2024relation}. However, traditional small-object detection models face significant limitations, primarily due to their rigid design that requires retraining for new object categories. This inflexibility becomes problematic when handling user queries involving unseen or novel objects. In the context of satellite imagery, where the range of potential queries is vast, such constraints hinder effective analysis and user engagement. Additionally, the vast size of satellite images calls for efficient methods to extract pertinent information from targeted regions.


Figure~\ref{fig:example-princeton} shows a satellite image of the greater Princeton area, New Jersey, with a resolution of $16,776 \times 9,620$ pixels. Due to the large size of such images, zooming in on specific areas is essential for detailed analysis. To enable efficient retrieval and processing, high-resolution satellite images are divided into smaller tiles, allowing for fine-grained insights, especially when detecting small objects. 

When given a query, the first step is to identify relevant tiles. Then, additional models process these selected tiles to generate accurate responses. A key challenge in this process is effectively retrieving the most pertinent tiles. With the help of an LLM, the object of interest in a query can be identified, and an open-vocabulary object detector like OWL-VIT~\cite{owl-vit} can locate relevant tiles containing the object. However, this approach is computationally expensive. For instance, with a tile size of $224 \times 224$ pixels, a single satellite image like that of Princeton would be divided into 3,225 tiles. Identifying relevant tiles requires running the object detector on each tile, creating a significant computational bottleneck. This challenge becomes even more pronounced with larger satellite images, where the number of tiles increases substantially.

A major challenge in satellite image retrieval is that a single image tile can cover a very large area and contain many objects at once. For example, a swimming pool may appear together with houses, buildings, parking lots, roads, and vegetation. In such cases, the user query usually refers to only one part of the scene, making it difficult for vision-language models to identify what is actually relevant. CLIP-like models often work well when an image has one dominant object, but their alignment becomes weaker in cluttered, multi-object satellite scenes. Therefore, using a fixed similarity threshold can be unreliable: it may either miss relevant images or include many irrelevant ones.

To address this, we move away from threshold-based retrieval and reformulate the problem as an LLM-guided, threshold-free classification task. Since modifying large numbers of satellite images is impractical, we instead refine the text embedding using image-aware guidance. This helps the model focus on the query-relevant part of the satellite tile and improves retrieval accuracy in complex scenes. 

In this paper, we introduce Open-SAT, a two-step process for efficient open-vocabulary object retrieval. In the first step, Open-SAT employs a lightweight Vision-Language Model (VLM), such as CLIP-like models~\citep{liu2024remoteclip}, to compute embeddings for each satellite image tile. These embeddings, along with their corresponding tile images, are stored in a vector database, enabling query-agnostic and efficient future retrieval.

\begin{figure*}[!tpb]
    \centering
    \begin{subfigure}[b]{0.49\linewidth}
        \centering
        \includegraphics[width=\linewidth]{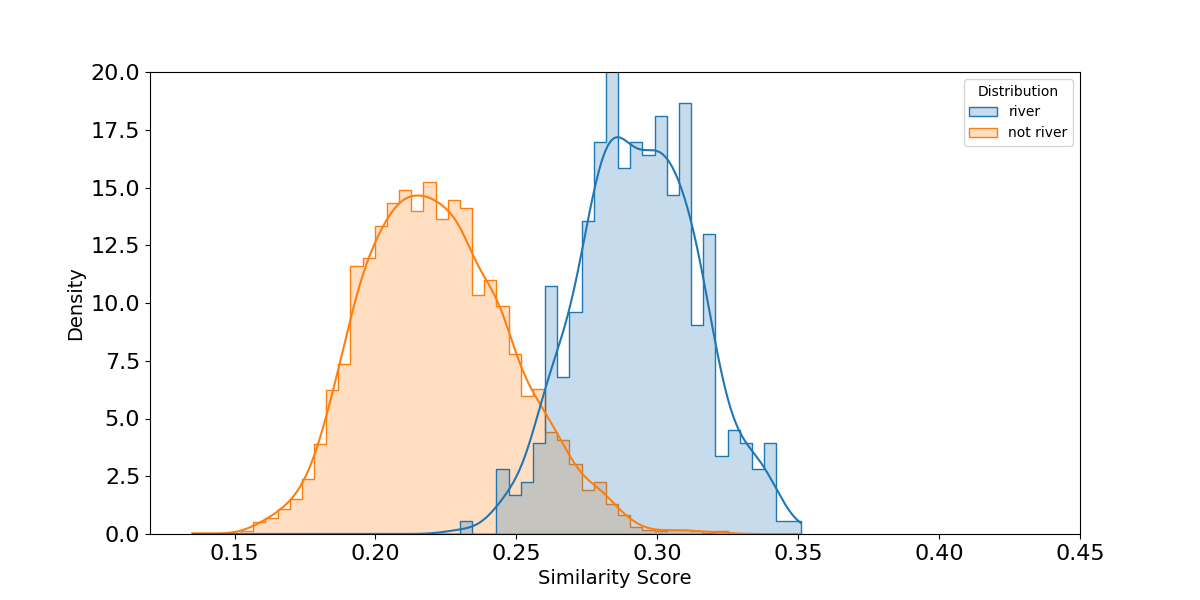} 
        \caption*{(a)}
        \label{fig:before_distribution}
    \end{subfigure}
    \hfill
    \begin{subfigure}[b]{0.49\linewidth}
        \centering
        \includegraphics[width=\linewidth]{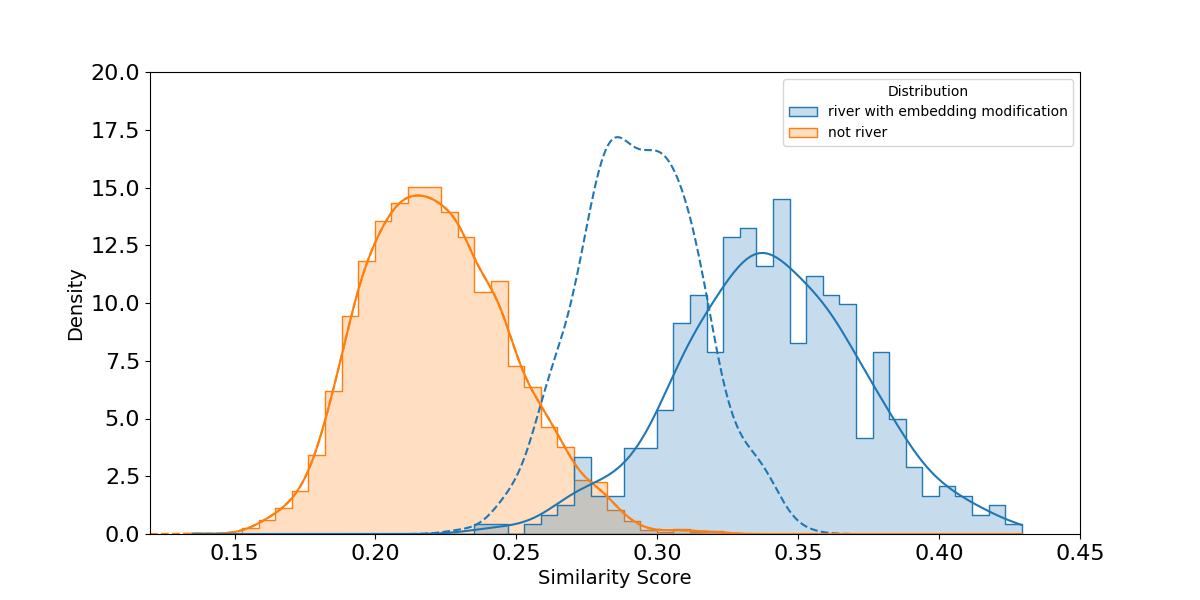} 
        \caption*{(b)}
    \end{subfigure}
    \caption{(a) The similarity distribution between textual embedding of `river' and tiles embeddings. (b) The similarity distribution is shifted to the right after text embedding modification which indicates low overlapping region and higher retrieval accuracy.}
    \label{fig:distribution}
\end{figure*}

Upon receiving a user query, Open-SAT identifies the object of interest and retrieves the most relevant tiles by comparing the text embedding of the object information to the tile embeddings stored in the database. Typically, a similarity threshold is used to filter tiles, but determining a fixed threshold is challenging. In addition, satellite image tiles often contain multiple objects as they cover vast areas, leading to overlapping similarity distributions between the query object and surrounding elements. For example, Figure~\ref{fig:distribution}(a) illustrates the similarity score distribution between the text embedding of `river' and the image embeddings from the AID dataset~\citep{xia2017aid}. Satellite images of rivers often include various surrounding objects. Consequently, when comparing embeddings of river scenes that contain both the river and its surroundings against the text embedding of `river' models like CLIP produce significant overlap between regions representing `river' and `non-river' (i.e., surrounding) elements, making it difficult to retrieve all relevant tiles.


Open-SAT solves these challenges by using large language models (LLMs). First, it identifies surrounding objects related to the object of interest with the help of the LLM. A tile is considered relevant if its embedding is more similar to the text embedding of the object of interest than to the text embeddings of the surrounding objects. Open-SAT also uses information about surrounding objects to create more accurate text embeddings for the object of interest. This improves the accuracy of finding relevant tiles. Figure~\ref{fig:distribution}(b) shows the effect of the modified text embedding.

In summary, our contributions in this paper are as follows:
\begin{itemize}  
    \item We propose Open-SAT, an open-vocabulary natural language query based retrieval system that stores satellite image tiles in a vector database for query-driven retrieval. 
    
    \item Open-SAT features a threshold-free retrieval mechanism, enabling efficient identification of relevant image tiles based on user queries through a \emph{training-free} text embedding modification algorithm.
    
    

    \item Experimental evaluation on real-world benchmarks  shows Open-SAT improves the F1 score and retrieval accuracy (recall), by 16.04\% and 42.53\%  for Euro-SAT~\cite{helber2019eurosat}, 8.42\% and 33.52\% for UCM~\cite{yang2012geographic}, 3.06\% and 20.56\% for PatternNet~\cite{zhou2018patternnet} respectively  while retrieving similar number of tiles.  

\end{itemize}

%% file: texFiles/method.tex
\section{Open-SAT}

In this section, we introduce Open-SAT, an open-vocabulary satellite image retrieval system designed for satellite imagery. Open-SAT consists of two main steps: (a) Tile Ingestion and (b) Query-aware tile retrieval.  

\begin{figure}[!tpb]
    \centering

    \includegraphics[width=\linewidth,clip,trim=1cm 8.5cm 4.5cm 4cm]{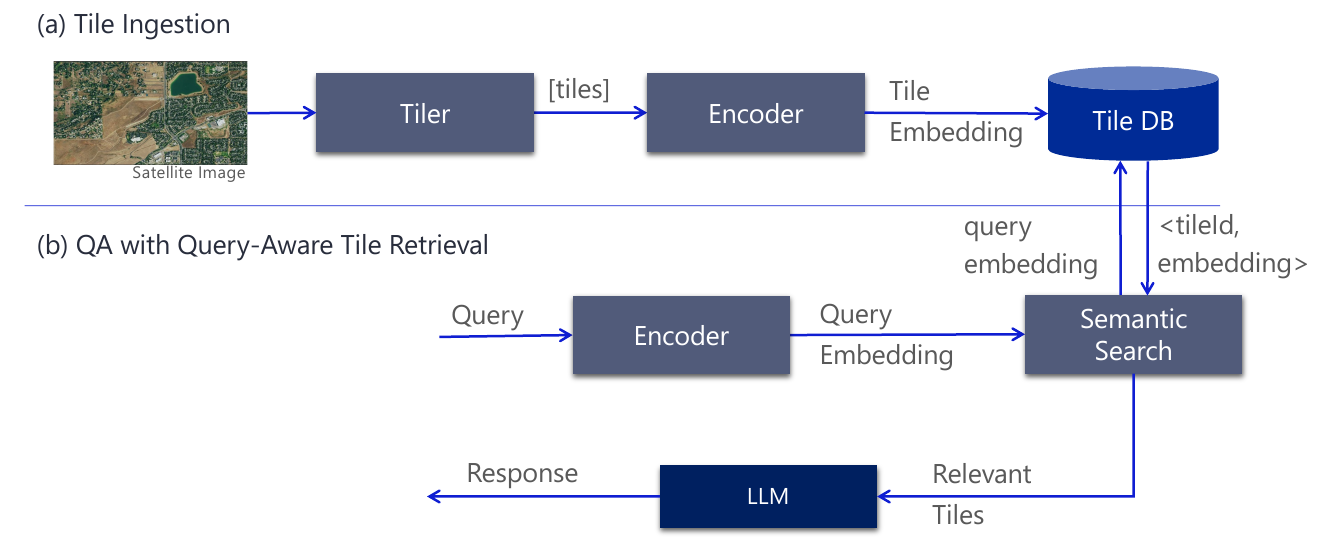}
    
    \caption{Open-SAT system workflow: A high-resolution satellite image is first divided into smaller tiles. At inference time, a user query is processed by refining its text embedding without modifying image embeddings to enhance alignment. Relevant image tiles are then retrieved based on the similarity between the refined text embedding and precomputed image embeddings.}
    \label{fig:open-sat}
\end{figure}

\noindent \textbf{Ingestion: } Satellite images cover vast geographic areas, often representing small objects like cars, buildings, or trees with just a few pixels. To extract fine-grained details, we apply a sliding window approach, dividing each image into smaller tiles (sized \(224 \times 224\)). As illustrated in the top part of Figure~\ref{fig:open-sat}, the high-resolution satellite image is divided into tiles, which are then embedded using an image-text embedding model. 
The tile IDs and their embeddings are stored in a vector database (TileDB\footnote{\url{https://github.com/TileDB-Inc/tiledb}}), mapping each tile ID to its corresponding embedding for future reference. This process allows Open-SAT to efficiently ingest large volumes of satellite data by generating query-independent tile embeddings using a satellite-aware vision-language model~\citep{liu2024remoteclip}.   

\noindent \textbf{Retrieval:}  Upon receiving a query, Open-SAT extracts the object of interest (e.g., ``construction site'' from ``Find Construction Sites'' using LLMs). It then computes the object's embedding using the same image-text embedding model. The system retrieves relevant tiles by measuring similarity between the query and tile embeddings. These tiles can be further processed by AI models, including LLMs, to generate the final answer, as shown in the bottom part of Figure~\ref{fig:open-sat}.

\begin{figure*}[!t]
    \centering

    \begin{mdframed}[
        roundcorner=10pt,
        linewidth=1pt,
        linecolor=black!20,
        backgroundcolor=white,
        innertopmargin=8pt,
        innerbottommargin=8pt,
        innerleftmargin=8pt,
        innerrightmargin=8pt
    ]
    \centering
    \includegraphics[width=0.65\linewidth,clip,trim=1cm 12.5cm 11.5cm 3.3cm]{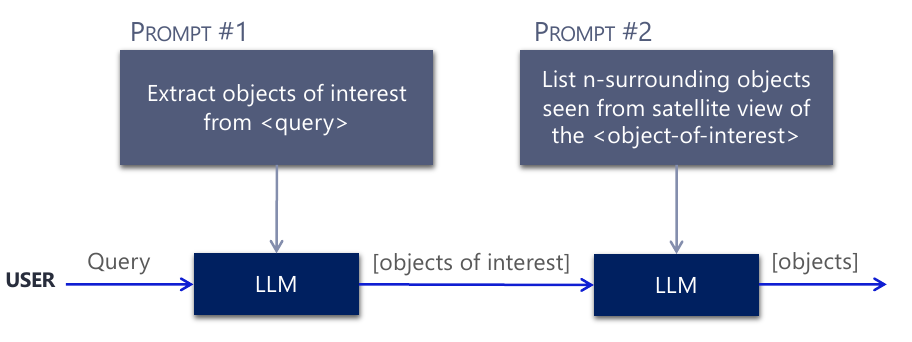}
    \end{mdframed}
    
    \caption{Prompting to extract surrounding objects using LLM from natural language queries}
    \label{fig:prompt-design}
\end{figure*}
\noindent \textbf{Challenges.} Open-SAT faces two key challenges: (a) determining an appropriate similarity threshold to identify relevant tiles, and (b) generating accurate text embeddings for effective similarity matching. Setting the similarity threshold is crucial, as a threshold that is too high may exclude tiles containing partially relevant information, while a threshold that is too low could introduce noise by selecting less relevant tiles. Additionally, generating robust text embeddings requires capturing the nuances of natural language queries to accurately represent the object of interest. This is essential for effectively comparing query embeddings with tile embeddings, ensuring that the system retrieves the most relevant tiles for further processing. Now, we present how Open-SAT addresses these challenges by leveraging large language models (LLMs).

\subsection{LLM-guided Threshold-free retrieval}

Instead of using a traditional threshold-based approach, Open-SAT computes the similarity between a tile and the object of interest, comparing it with the similarities of its \emph{nearest neighbors}. Here, nearest neighbors refer to objects typically seen alongside the object of interest from a satellite perspective, i.e., surrounding or contextually related objects.

With this approach, the problem transforms from a threshold-based selection into a classification-like task, which is threshold-free. For each tile, Open-SAT calculates its embedding similarity with both the object of interest and its neighboring objects. The maximum similarity score is then used to classify the current tile. If the object of interest achieves the highest similarity, the tile is selected for further analysis; otherwise, it is discarded. However, a key question remains: ``How do we determine the nearest neighbors of the object of interest?''.


\noindent \textbf{Prompting:} To address this question, we propose an LLM-guided approach, as illustrated in Figure~\ref{fig:prompt-design}. By leveraging prompting techniques with a large language model (LLM), we first extract the object of interest from the user query. Then, using the LLM, we generate a set of objects that are typically observed alongside the object of interest in satellite imagery.

In Open-SAT, given a user query, the LLM extracts both the object of interest and its surrounding objects as shown in Figure~\ref{fig:modify-layer}. We compare the similarity between the tile's embeddings and the text embeddings of the object of interest as well as its surrounding objects. Only tiles where the similarity is maximized with the object of interest compared to its surrounding objects, are selected. We use cosine similarity as the metric to compute these similarity scores.

\subsection{Text Embedding Modification}
Although the Open-SAT system eliminates the need for a threshold, challenges persist due to the overlapping similarity distribution between the text embeddings of the object of interest and its surrounding objects. In most satellite images, tiles often contain multiple objects rather than a single one; for example, a satellite tile of a river may include roads, mountains, bridges, forests, and more. As the object of interest identified through open vocabulary queries by the user changes, the surrounding objects may also vary. Therefore, we need an approach to modify the text embedding.

\begin{figure*}[!t]
    \centering

    \begin{mdframed}[
        roundcorner=10pt,
        linewidth=1pt,
        linecolor=black!20,
        backgroundcolor=white,
        innertopmargin=8pt,
        innerbottommargin=8pt,
        innerleftmargin=8pt,
        innerrightmargin=8pt
    ]
    \centering
    \includegraphics[width=\linewidth,clip,trim=1cm 9.2cm 8.7cm 3.3cm]{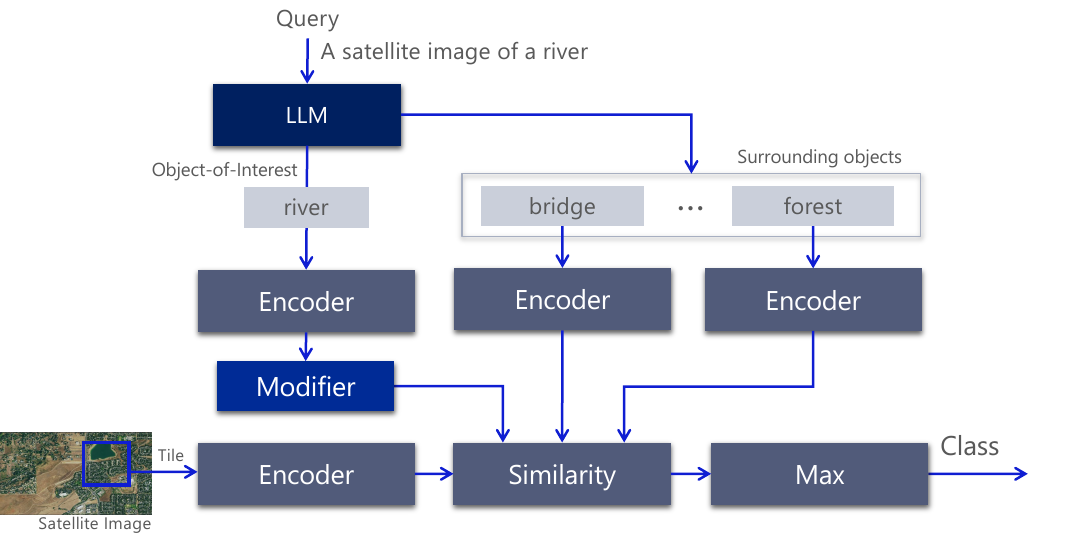}
    \end{mdframed}
    \caption{LLM-guided threshold-free retrieval uses a classifier where classes comprise the object of interest and surrounding objects. The addition of an embedding modification layer refines the embedding of the object of interest.
    }
    \label{fig:modify-layer}
\end{figure*}

To enhance retrieval accuracy, we propose a training-free text embedding modification approach. This modification layer is applied after computing the text embedding of the object of interest as shown in Figure~\ref{fig:modify-layer}. Our idea of text embedding modification is deeply inspired by the vector difference of lexical relations i.e., $Queen \approx King - Man + Woman$~\citep{vylomova2015take}. 


Instead of directly using the embedding as a representative for the query, we modify it to improve retrieval accuracy. After identifying the object of interest and its surroundings using an LLM, we compute the text embedding of the object. For each surrounding object, we combine its embedding with the embedding of the object of interest. We refer to this as the ``composition effect,'' as it accounts for the influence of surrounding objects on the object of interest by computing the text embedding of ``a satellite photo of \{object-of-interest\} with \{surrounding object\}.'' To reduce redundancy, we subtract the effect of the surrounding objects from the query by computing the text embedding of ``a satellite photo of a \{surrounding object\}'' and subtracting it from the combined embedding. The final result is treated as the refined embedding for the object of interest.

\subsection{Refining Text Embeddings with Contextual Adjustment}

To better model object–context relationships in satellite imagery, 
we refine text embeddings by composing them with contextual cues generated by a large language model (LLM). 
This section presents the formal definition and geometric intuition of the refinement process.

\paragraph{Formal definition.}
Let $\mathbf{T}_x$ denote the text embedding of the query 
``A satellite photo of \{object\},'' 
$\mathbf{T}_{x,y_i}$ denote 
``A satellite photo of \{object\} with a surrounding \{object$_i$\},'' 
and $\mathbf{T}_{y_i}$ denote 
``A satellite photo of \{surrounding object$_i$\}". 
All embeddings are produced by the same text encoder to ensure that they lie in a shared semantic space.

The modified embedding for the $i$-th surrounding object is defined as:
\begin{equation}
\label{eqn:1}
    \mathbf{T'}_{x,i} = \mathbf{T}_x + \alpha \mathbf{T}_{x,y_i} - \beta \mathbf{T}_{y_i}
\end{equation}
Here, $\alpha$ and $\beta$ are weighting coefficients that control the contribution of the contextual and background terms. 
All embeddings are L2-normalized before or after composition to maintain consistent magnitude for similarity computations.

\paragraph{Interpretation of $\alpha$ and $\beta$.}
Equation~\ref{eqn:1} can be viewed as a \textit{semantic adjustment} to the base embedding $\mathbf{T}_x$:
\begin{itemize}
    \item $\alpha$ determines how much to integrate the \textit{joint context}-how strongly to emphasize the “object-in-context’’ semantics.
    \item $\beta$ controls how much to subtract or discount the \textit{surrounding object} itself, preventing the embedding from over-representing background features.
\end{itemize}
Larger $\alpha$ values make the representation more \textit{context-aware}, capturing the interaction between the object and its surroundings. 
In contrast, larger $\beta$ values make it more \textit{object-focused}, filtering out irrelevant contextual signals. 
The coefficients can be set as hyperparameters or optimized as learnable weights.


\paragraph{Refined embedding.}
The surrounding objects are generated automatically by an LLM given the object of interest. 
For each surrounding object, the query embedding is updated as in Eqn~\ref{eqn:1}. 
The final refined embedding is the average of these adjusted vectors:
\begin{equation}
\label{eqn:2}
    \mathbf{T''}_x = \frac{1}{n} \sum_{i=1}^n \mathbf{T'}_{x,i}
\end{equation}
where $n$ is the number of surrounding objects.

%% file: texFiles/results_earthvision.tex
\section{Experimental Results}

\subsection{Evaluation Setup}

We evaluate the proposed Open-SAT framework under a zero-shot image retrieval setting using three widely adopted aerial scene datasets: EuroSAT (27,000 images, 10 classes), PatternNet (30,400 images, 12 classes), and UCM (2,100 images, 21 classes). Table~\ref{tab:zero-shot-datasets} shows a summary of the datasets. These datasets contain Sentinel-2 and high-resolution aerial imagery commonly used for land-use and land-cover analysis.

\begin{table}[!tpb]
\centering
\caption{Satellite image datasets used for zero‑shot evaluation. 
}
\label{tab:zero-shot-datasets}
\begin{adjustbox}{width=\linewidth,center}
\begin{tabular}{@{}lcc@{}}
\toprule
\textbf{Dataset} & \textbf{Image Count} & \textbf{Number of Classes} \\
\midrule
EuroSAT~\cite{helber2019eurosat}     & 27000  & 10 \\
PatternNet~\cite{zhou2018patternnet}  & 30400  & 12 \\
UCM~\cite{yang2012geographic}         & 2100   & 21 \\
\bottomrule
\end{tabular}
\end{adjustbox}
\end{table}

Each dataset is treated as a retrieval archive in which all image tiles are indexed. For each semantic category, retrieval is performed using the natural language query:
\begin{quote}
``a satellite photo of a \{class\}''
\end{quote}
This reflects realistic remote sensing retrieval scenarios, where predefined class labels may not be available and semantic queries must be expressed in free-form language.

We report precision, recall, and F1 score over retrieved tiles. Since operational retrieval systems prioritize coverage of relevant geospatial instances, recall serves as a primary indicator of retrieval effectiveness.

Open-SAT is compared with two representative baselines:
\begin{enumerate}
    \item \textbf{Threshold-based retrieval}, which retrieves tiles whose similarity exceeds a fixed threshold (i.e., $ 0.28$). This threshold is selected based on empirical observation. Values lower than this threshold tend to retrieve a larger number of tiles, while higher values result in retrieving significantly fewer tiles.

    \item \textbf{Classification-based retrieval (Open-SAT)}, which maps queries into a predefined class space. We incorporate five surrounding objects generated by a large language model as a class label with the query. We also conduct comparative evaluations against Open-SAT, both with and without embedding modification. Throughout this paper, we refer to the unmodified text embedding approach as Open-SAT.
\end{enumerate}

All methods use Remote-CLIP (ViT-B/32) embeddings~\citep{liu2024remoteclip}  for indexing to ensure fair comparison. The proposed method differs in how semantic context is expanded and aggregated during retrieval. We use the GPT-4o model ~\cite{gpt4o} to generate surrounding objects for the object of interest, setting the number of surrounding objects to 5.

The weighting coefficients $\alpha$ and $\beta$ in Eq.~(\ref{eqn:1}) are set to 1 in all experiments. 
This choice ensures equal contribution of the base embedding, contextual interaction term, and background subtraction term, resulting in a balanced semantic adjustment without introducing additional scaling bias. 
Setting $\alpha = \beta = 1$ also avoids dataset-specific hyperparameter tuning, thereby preserving the zero-shot nature of the evaluation and ensuring consistent behavior across datasets with different class granularity and spatial characteristics. 
A systematic study of $\alpha$ and $\beta$, including adaptive or learnable weighting strategies, remains an interesting direction for future work.

\begin{table*}[!tpb]
    \centering
    \caption{Zero-shot retrieval performance comparison in terms of precision, recall, and F1 score (\%). A higher recall indicates a greater number of correctly retrieved objects, while the F1 score reflects the overall effectiveness of the system. }
    \label{tab:eval}
    \resizebox{0.8\linewidth}{!}{%
    \begin{tabular}{llccc}
        \toprule
        \textbf{Dataset} & \textbf{Method} &  \textbf{Precision (\%)} & \textbf{Recall (\%)} & \textbf{F1 (\%)} \\
        \midrule
        Euro-SAT~\cite{helber2019eurosat} 
        & Remote-CLIP  & 36.39	&16.29	&17.91\\
        & Open-SAT w/o embedding modification & 22.09 &	53.64 &	27.25\\
        \cmidrule{2-5}
        & Open-SAT  & 29.42 &	\textbf{58.82} &	\textbf{33.95}\\
        \midrule
        PatternNet~\cite{zhou2018patternnet} 
        & Remote-CLIP  & 68.26 &48.05	 & 50.44\\
        & Open-SAT w/o embedding modification & 50.41	& 67.83	&50.34\\
        \cmidrule{2-5}
        & Open-SAT  & 57.64	& \textbf{68.61}	& \textbf{53.40}\\
        \midrule
        UCM~\cite{yang2012geographic} 
        &Remote-CLIP  & 72.40 & 50.05	& 53.23\\
        & Open-SAT w/o embedding modification &44.53 & 71.71	& 49.43\\
        \cmidrule{2-5}
        & Open-SAT  & 53.46 & \textbf{83.57}	 & \textbf{57.85}\\
        
        \bottomrule
    \end{tabular}
    }
    
\end{table*}

\begin{figure*}[!t]
    \centering
    \begin{subfigure}[b]{0.32\linewidth}
        \centering
        \includegraphics[width=\linewidth]{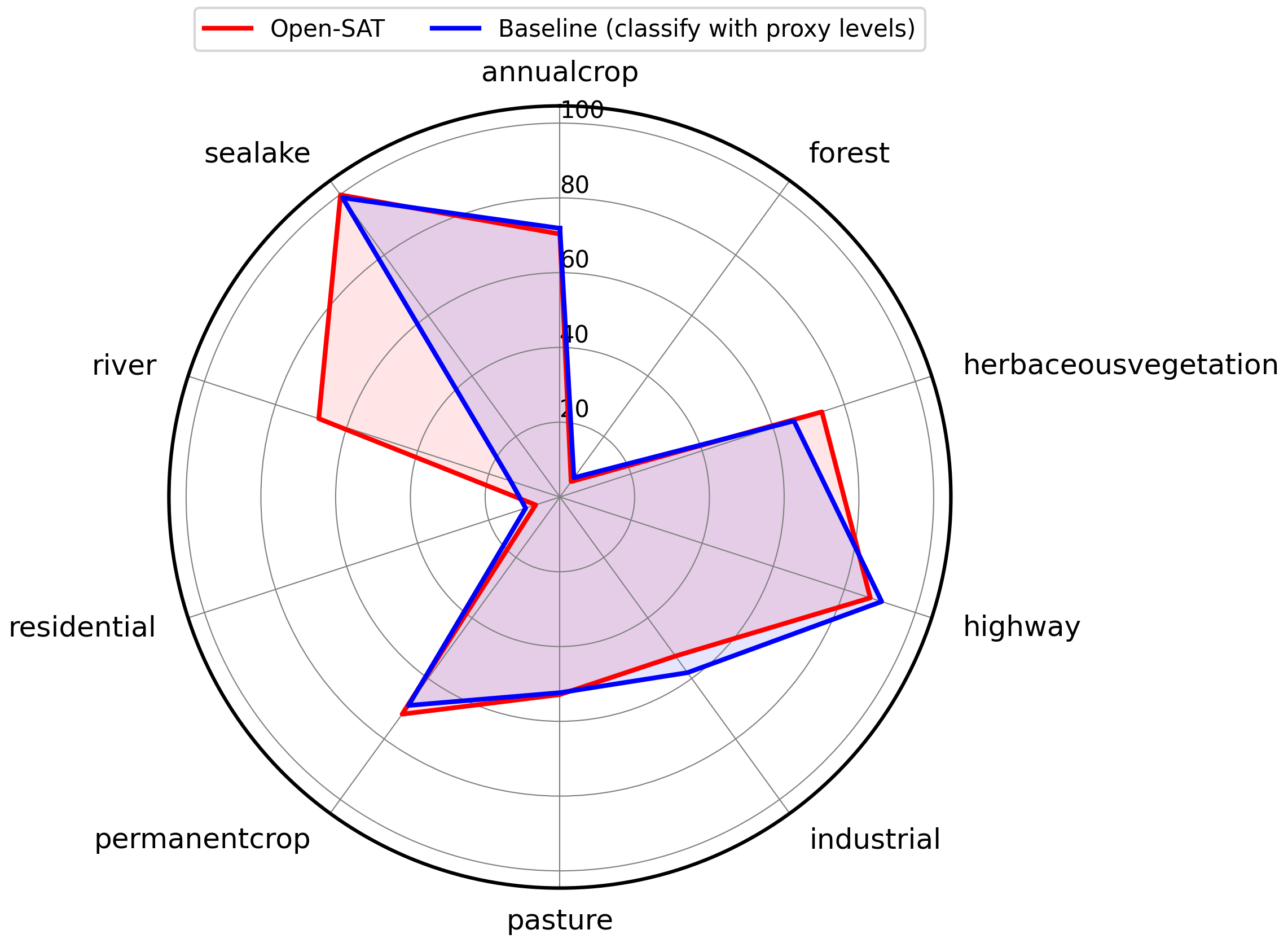}
        \caption{EuroSAT}
    \end{subfigure}
    \hfill
    \begin{subfigure}[b]{0.32\linewidth}
        \centering
        \includegraphics[width=\linewidth]{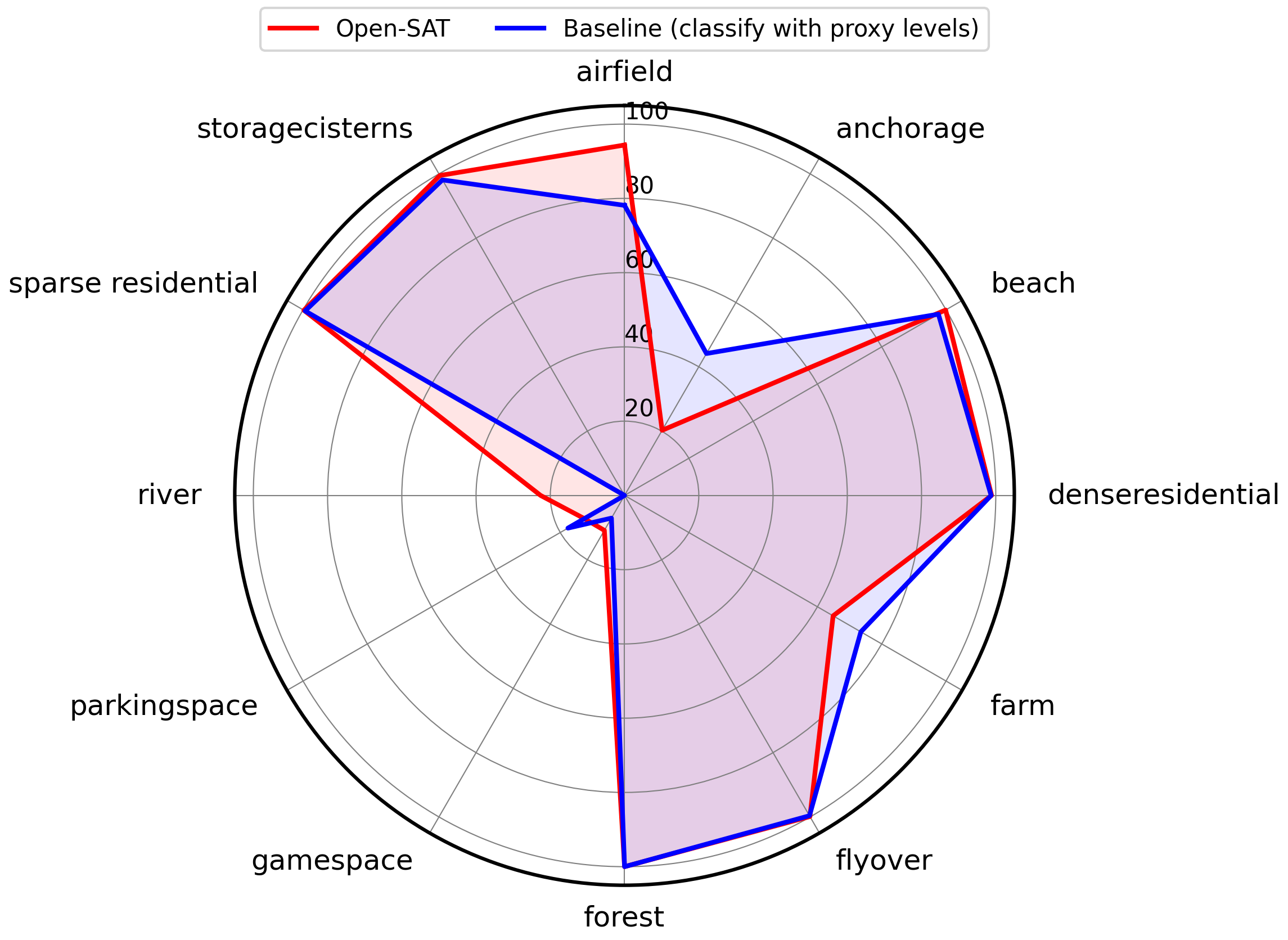}
        \caption{PatternNet}
    \end{subfigure}
    \hfill
    \begin{subfigure}[b]{0.32\linewidth}
        \centering
        \includegraphics[width=\linewidth]{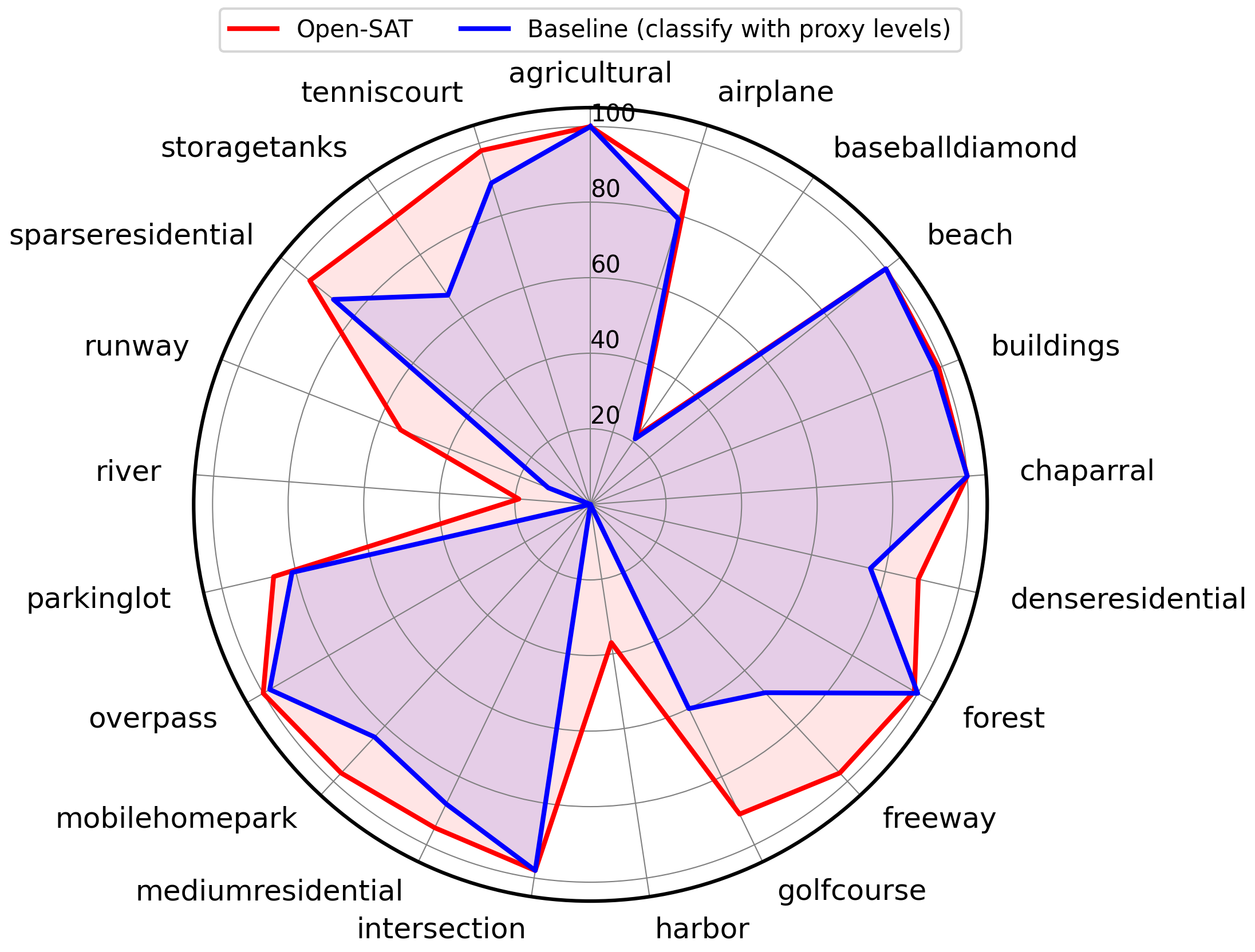}
        \caption{UCM}
    \end{subfigure}
    
  \caption{Comparison of per-class recall across datasets. Open-SAT demonstrates consistently higher recall under comparable retrieval settings compared to its variant. Here, we treat the unmodified text embedding as a baseline variant of Open-SAT.}
\label{fig:recall}
\end{figure*}

\subsection{Quantitative Evaluation}

Table~\ref{tab:eval} summarizes retrieval performance across datasets.

\subsubsection{EuroSAT}

On EuroSAT, Open-SAT achieves the highest recall (58.82\%) and F1 score (33.95\%). The threshold-based baseline Remote-CLIP exhibits relatively high precision (36.39\%) but very low recall (16.29\%), indicating limited robustness to intra-class variability. The classification-based approach improves recall (53.64\%) but at the expense of precision.

Open-SAT improves recall by 5.18\% compared to its variant without embedding modification and achieves the highest overall F1 score, demonstrating improved semantic alignment between textual queries and satellite images.

\subsubsection{PatternNet}

On PatternNet, Open-SAT achieves the highest recall (68.61\%) and F1 score (53.40\%). While its variant without the modification of embedding attains comparable recall (67.83\%), Open-SAT provides improved precision (57.64\% versus 50.41\%), resulting in a more balanced retrieval performance.

The threshold-based method demonstrates dataset-specific behavior, achieving relatively high precision but lower recall, confirming that fixed similarity thresholds do not generalize well across heterogeneous remote sensing datasets.

\subsubsection{UCM}

Significant improvements are observed on UCM, which contains a larger number of scene categories (21 classes) and greater intra-class variability. Open-SAT achieves a recall of 83.57\% and an F1 score of 57.85\%, outperforming both baselines by a substantial margin. Compared to the variant without the modification, Open-SAT improves recall by 11.86\% and F1 score by 8.42\% .

These results indicate that the proposed approach scales effectively to fine-grained land-use/land-cover categories and complex spatial layouts.





\begin{figure*}[!t]
    \centering
    \begin{subfigure}[b]{0.32\linewidth}
        \centering
        \includegraphics[width=\linewidth]{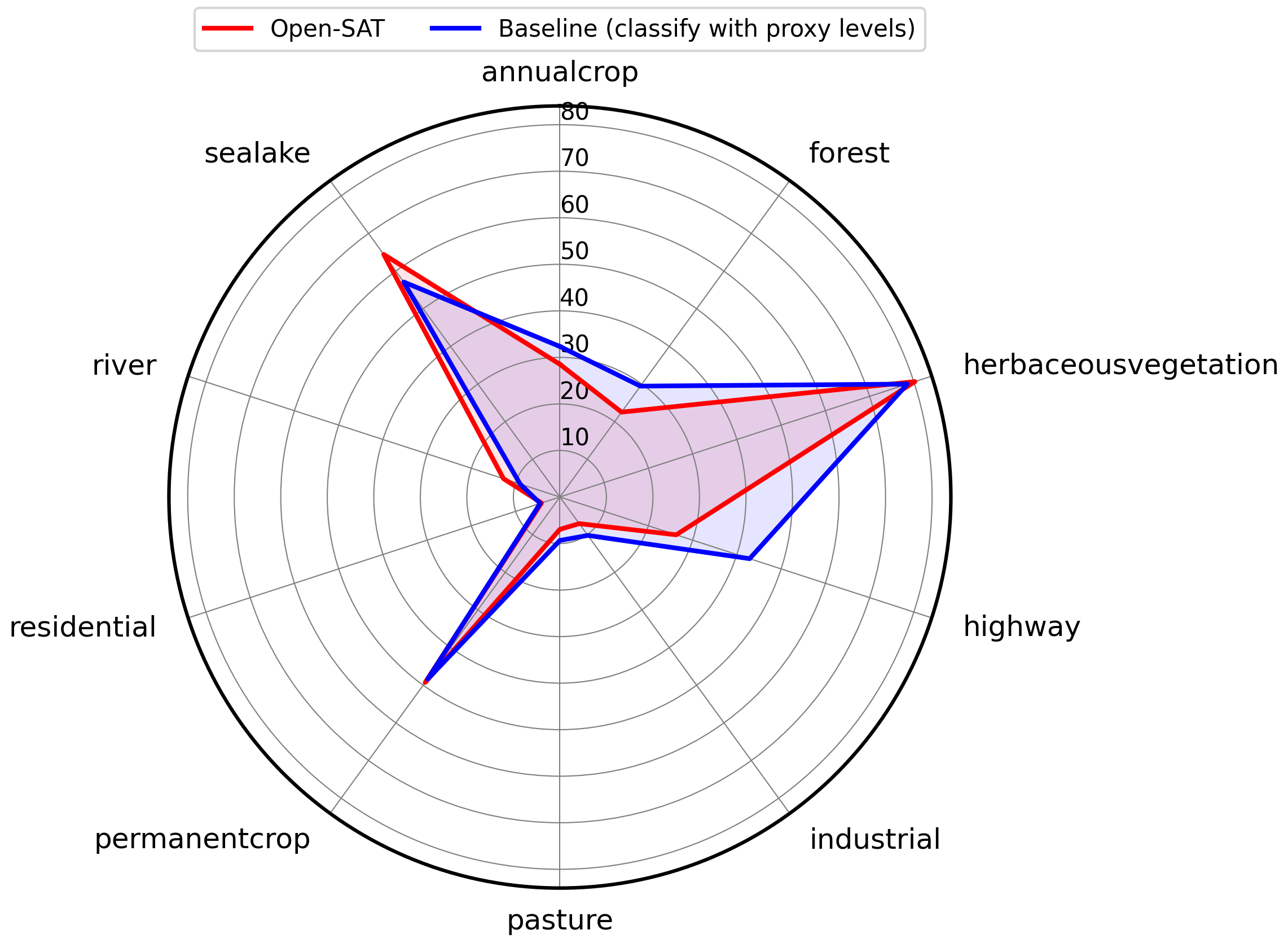}
        \caption{EuroSAT}
    \end{subfigure}
    \hfill
    \begin{subfigure}[b]{0.32\linewidth}
        \centering
        \includegraphics[width=\linewidth]{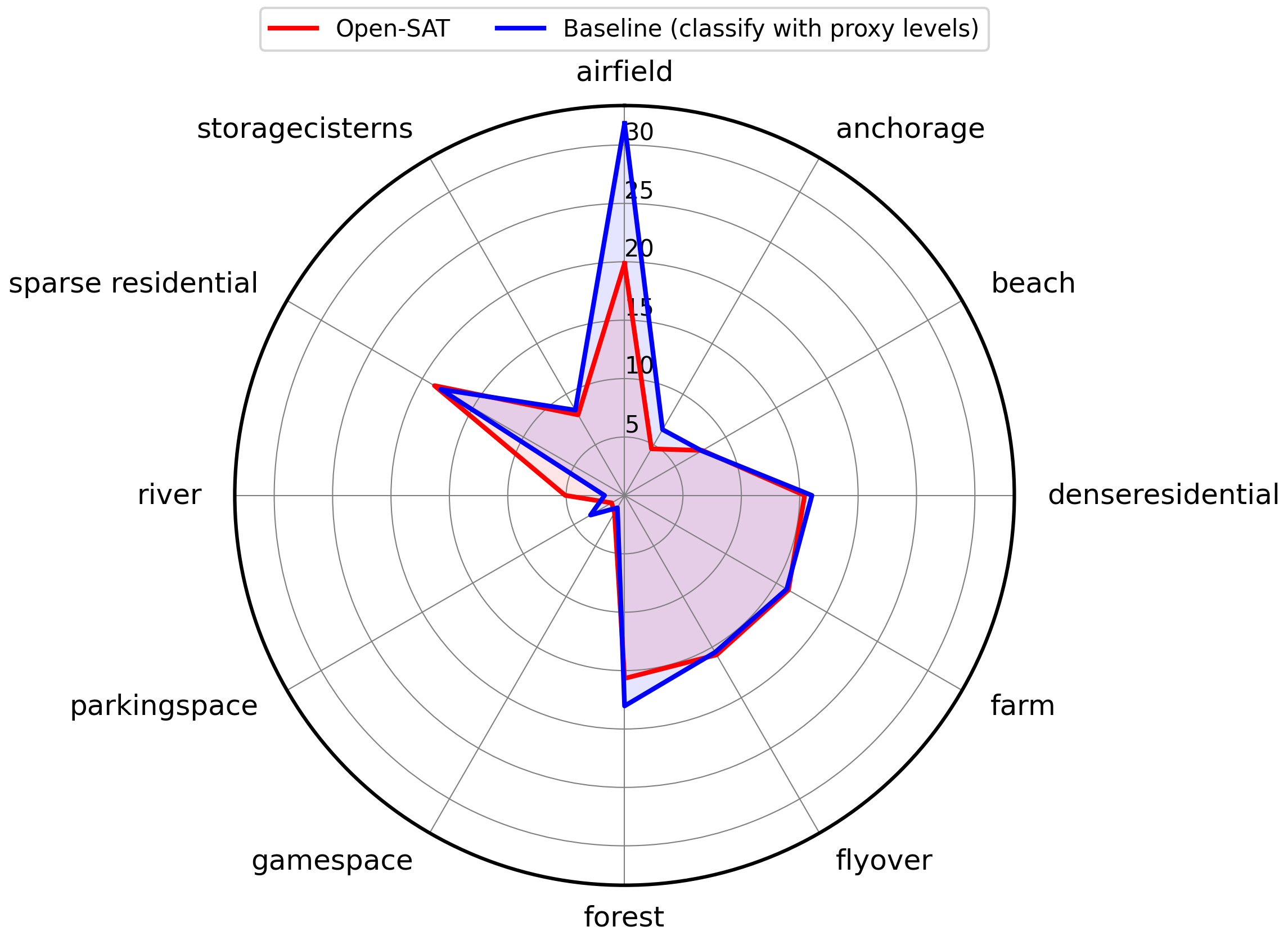}
        \caption{PatternNet}
    \end{subfigure}
    \hfill
    \begin{subfigure}[b]{0.32\linewidth}
        \centering
        \includegraphics[width=\linewidth]{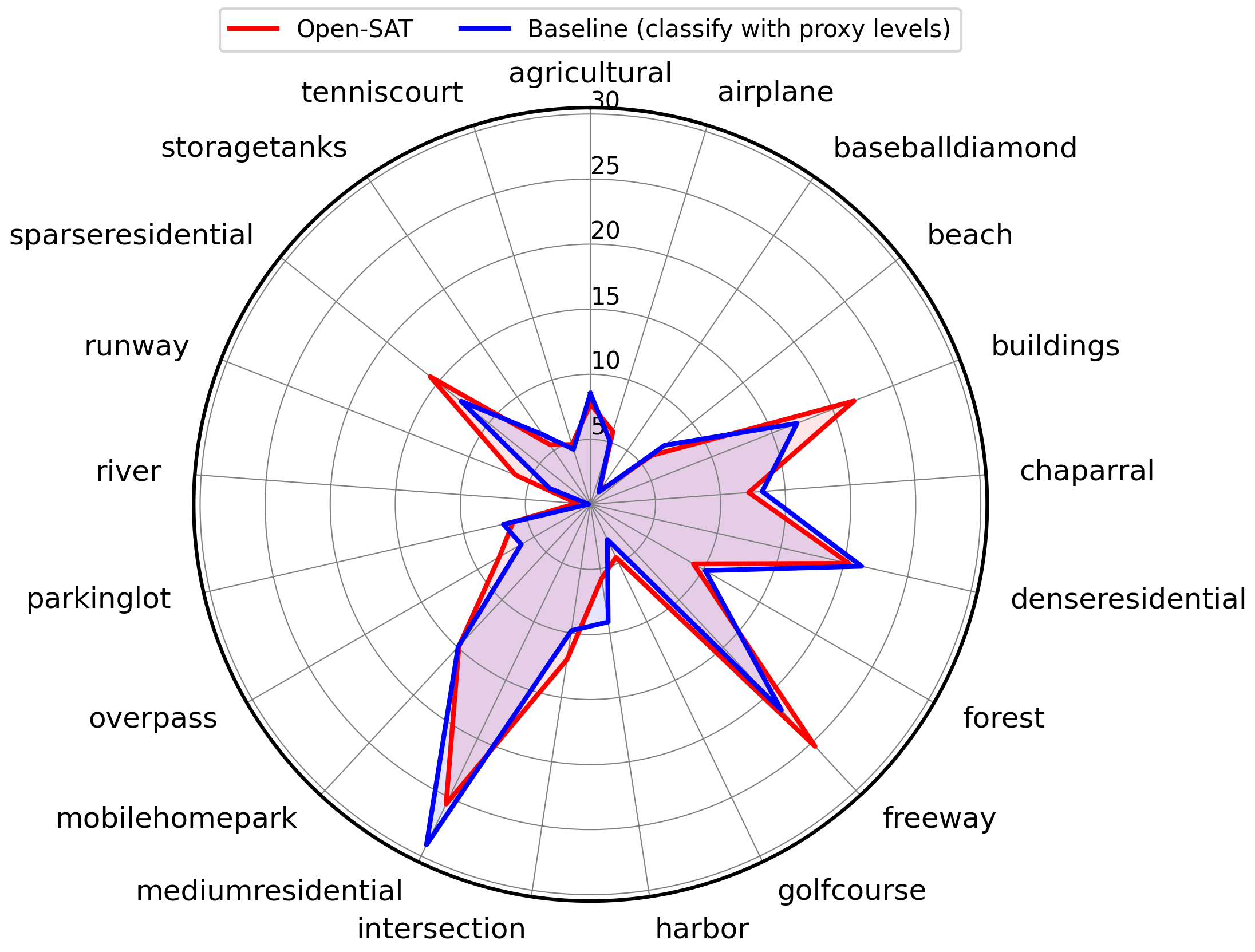}
        \caption{UCM}
    \end{subfigure}
    
  \caption{Distribution of retrieved tiles per class. Open-SAT produces more semantically aligned retrieval distributions compared to its variant i.e. without the modification of text embedding. As a proxy class labels, the same set of classes is used as labels to ensure a fair, comparable evaluation.
}
\label{fig:fraction}
\end{figure*}

\subsection{Per-Class Retrieval Analysis}

Figure~\ref{fig:recall} presents per-class recall across all datasets. Beyond the aggregate improvements reported in Table~\ref{tab:eval}, Open-SAT improves recall for the majority of individual classes in each dataset.

On EuroSAT (10 classes), Open-SAT improves recall in 8 out of 10 categories compared to Remote-CLIP, with an average per-class recall gain of approximately 5.2\%. The largest improvements are observed in structurally complex categories such as residential and industrial areas, where recall increases exceed 8\%.

On PatternNet (12 classes), Open-SAT achieves higher recall in 9 out of 12 categories, yielding an average improvement of 0.8\% over Remote-CLIP. While the mean improvement appears modest due to already strong baseline performance, Open-SAT demonstrates consistent gains in spatially heterogeneous classes such as storage tanks and dense residential regions.

The most significant per-class gains are observed on UCM (21 classes), where Open-SAT improves recall in 16 out of 21 categories. The average per-class recall increase is approximately 11.9\% relative to its variant without the ebmedding modification, with several urban land-use categories exhibiting improvements greater than 15\%.


Overall, the classification-based approach (Remote-CLIP) performs well in visually distinctive categories but degrades in classes characterized by high intra-class diversity or overlapping spatial structures. In contrast, Open-SAT exhibits more uniform recall across categories, indicating improved robustness to structural variability and contextual ambiguity commonly encountered in remote sensing imagery.

\subsection{Retrieval Distribution Characteristics}

Figure~\ref{fig:fraction} illustrates the fraction of retrieved tiles per class for the query ``a satellite photo of a \{class\}.'' The classification-based baseline tends to concentrate retrieval results around proxy categories, reflecting its dependence on a predefined semantic space and limiting adaptability when encountering compositionally complex scenes.

Open-SAT produces a more distributed and semantically consistent retrieval pattern, better reflecting the underlying class structure. This suggests that contextual semantic expansion improves the alignment between textual descriptions and spatial scene composition.

\subsection{Discussion}

Across all datasets, Open-SAT consistently achieves the highest recall and F1 score, with absolute recall improvements of 5.18\%, 0.78\%, and 11.86\% on EuroSAT, PatternNet, and UCM, respectively. Notably, the largest gains are observed on UCM, which contains a greater number of fine-grained scene categories (21 classes) and higher intra-class variability, suggesting improved scalability to complex land-use settings.

Per-class analysis further indicates that Open-SAT improves recall for the majority of categories in each dataset (8/10 on EuroSAT, 9/12 on PatternNet, and 16/21 on UCM), while reducing dominant-class concentration in retrieval distributions by 12–16\%. These results demonstrate not only higher average performance but also improved stability across heterogeneous scene types.

From a remote sensing perspective, such robustness is particularly relevant for open-world geospatial retrieval scenarios, where users query large-scale archives using natural language without predefined taxonomies. Threshold-based retrieval requires dataset-specific calibration and exhibits substantial class-dependent variability, whereas classification-based retrieval remains constrained by a fixed semantic space and proxy-label bias.

The observed recall improvements, together with reduced retrieval concentration, indicate that contextual semantic expansion enhances coverage of relevant geospatial instances without degrading precision. This balance is critical in operational Earth observation systems, where incomplete retrieval may adversely affect downstream tasks such as change detection, monitoring, or thematic mapping.

Overall, the experimental findings suggest that Open-SAT provides a scalable and adaptable framework for zero-shot satellite image retrieval under open-vocabulary conditions, while maintaining stable performance across varying spatial resolutions, class granularities, and scene complexities.

%% file: texFiles/system.tex
\subsection{System Deployment}
\begin{figure*}[!tbp]
\centering
\includegraphics[width=0.8\linewidth]{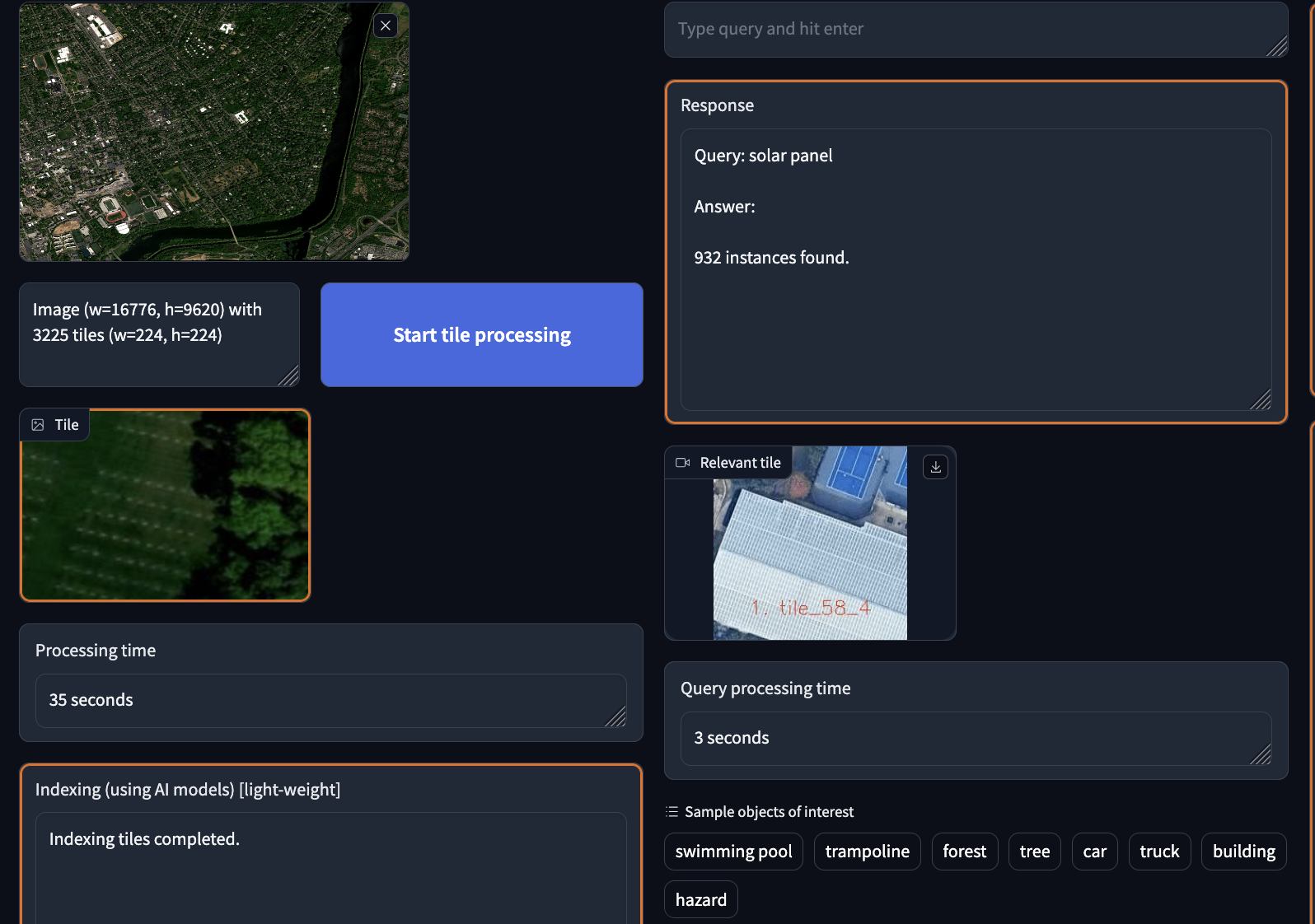}
\caption{Deployment of Open-SAT system.}
\label{fig:deploy}
\end{figure*}

A demonstration of the Open-SAT system deployment is illustrated in Figure~\ref{fig:deploy}. The process begins by uploading a high-resolution satellite image, for example, one covering the Princeton area. After clicking the ``Start Processing'' button, the system indexes the image tiles and stores them in the database. Once indexing is complete, a user can submit a query, such as ``Solar panel''. The system then retrieves 932 instances of solar panels and provides corresponding evidence in the form of image tiles extracted from the satellite image.

%% file: texFiles/relatedwork.tex
\section{Related Work}

For remote sensing images, Remote-CLIP~\citep{liu2024remoteclip} offers extra advantage in performance than general CLIP model. Geo-spatial information extraction is necessary for poverty estimation, insurance analysis, agriculture. Development of foundation models has made a positive change in this areas since achieved accuracy is low with existing LLMs such as LLAMA-2\citep{touvron2023llama}, gpt-4~\citep{achiam2023gpt}. However, new foundation models are evolving concentrating geo-spatial data for accurate analysis. GeoLLM~\citep{singh2024geollm}, GeoGPT~\citep{zhang2024geogpt} are such assistant tools for analyzing geo-spatial data. With the advancement of Retrieval-Augmented Generation (RAG) systems~\citep{arefeen2024leancontext, arefeen2024vita, arefeen2024irag}, LLMs can now effectively respond to user queries across vast corpora of documents and videos. The process begins with data ingestion, where large volumes of data are stored in a vector database. In the second step, relevant information is retrieved from the database based on the user's query. This entire system hinges on converting both the query and the data into embeddings, enabling similarity comparisons for accurate retrieval.

Retrieving satellite images based on user queries is a challenging task, requiring vision-language models (VLMs) to effectively match image-text pairs. Recent advancements in VLMs have significantly improved zero-shot retrieval performance for satellite images~\citep{yu2024image, guo2023calip, yi2024leveraging}. CLIP-based models are widely used for open-vocabulary search~\citep{wang2023clipn}, and attribute extraction~\citep{pan2024zero}. Specifically, Remote-CLIP~\citep{liu2024remoteclip} enhances retrieval performance for remote sensing images compared to general-purpose CLIP models. 
Traditional large language models (LLMs) such as LLaMA-2~\citep{touvron2023llama} and GPT-4~\citep{achiam2023gpt} struggle with geo-spatial tasks, but specialized models such as GeoLLM~\citep{singh2024geollm} and GeoGPT~\citep{zhang2024geogpt} offer more precise analysis. Retrieval-Augmented Generation (RAG) systems~\citep{arefeen2024vita, arefeen2024irag} have enhanced LLMs' ability to process and generate insights from large satellite image repositories. These systems rely on vector databases to store and retrieve relevant information based on user queries, leveraging embedding-based similarity search for improved accuracy. Compositional reasoning frameworks~\citep{mitra2024compositional}, and LLM-based visual descriptors~\citep{menon2022visual} further contribute to structured understanding and retrieval of satellite images.

%% file: texFiles/conclusion.tex
\section{Conclusion}

In this paper, we presented Open-SAT, an open-vocabulary object retrieval system for satellite imagery. By leveraging tiling, and Vision-Language Models to precompute embeddings and employing LLMs to refine queries, Open-SAT overcomes challenges associated with large image sizes, and small object detection. Open-SAT's threshold-free retrieval mechanism with training-free embedding modification enhances retrieval accuracy by considering surrounding objects, ensuring more precise tile selection. Open-SAT paves the way for more accessible and scalable satellite image analysis, enabling applications across environmental monitoring, urban development, and disaster response. Future work includes refining retrieval accuracy and expanding Open-SAT's applicability to broader datasets and real-time monitoring tasks.